\documentclass[letterpaper]{article} % DO NOT CHANGE THIS
\usepackage{aaai2026}  % DO NOT CHANGE THIS
\usepackage{times}  % DO NOT CHANGE THIS
\usepackage{helvet}  % DO NOT CHANGE THIS
\usepackage{courier}  % DO NOT CHANGE THIS
\usepackage[hyphens]{url}  % DO NOT CHANGE THIS
\usepackage{graphicx} % DO NOT CHANGE THIS
\usepackage{natbib}  % DO NOT CHANGE THIS AND DO NOT ADD ANY OPTIONS TO IT
\usepackage{caption} % DO NOT CHANGE THIS AND DO NOT ADD ANY OPTIONS TO IT
\usepackage{algorithm}
\usepackage{algorithmic}

\usepackage{amsmath}
\usepackage[version=4]{mhchem}
\usepackage{amssymb}

\usepackage{newfloat}
\usepackage{listings}
\DeclareCaptionStyle{ruled}{labelfont=normalfont,labelsep=colon,strut=off} % DO NOT CHANGE THIS
\floatstyle{ruled}
\newfloat{listing}{tb}{lst}{}
\floatname{listing}{Listing}
\title{Life, Machine Learning, and the Search for Habitability: Predicting Biosignature Fluxes for the Habitable Worlds Observatory}
\author{
Mark Moussa\textsuperscript{\rm 1},
Amber V. Young\textsuperscript{\rm 1},
Brianna Isola\textsuperscript{\rm 1,\rm 3},
Vasuda Trehan\textsuperscript{\rm 1,\rm 4},
Michael D. Himes\textsuperscript{\rm 1,\rm 5},
Nicholas Wogan\textsuperscript{\rm 2},
Giada Arney\textsuperscript{\rm 1}
}
\affiliations{
\textsuperscript{\rm 1}NASA Goddard Space Flight Center, Greenbelt, MD 20771, USA\\
\textsuperscript{\rm 2}NASA Ames Research Center, Moffett Field, CA 94035, USA\\
\textsuperscript{\rm 3}University of New Hampshire, Durham, NH 03824, USA\\
\textsuperscript{\rm 4}University at Albany (SUNY), Albany, NY 12222, USA\\
\textsuperscript{\rm 5}Morgan State University, Baltimore, MD 21251, USA

}

\begin{document}

\maketitle

\begin{abstract}
Future direct-imaging flagship missions, such as NASA's Habitable Worlds Observatory (HWO), face critical decisions in prioritizing observations due to extremely stringent time and resource constraints. In this paper, we introduce two advanced machine-learning architectures tailored for predicting biosignature species fluxes from exoplanetary reflected-light spectra: a Bayesian Convolutional Neural Network (BCNN) and our novel model architecture, the Spectral Query Adaptive Transformer (SQuAT). The BCNN robustly quantifies both epistemic and aleatoric uncertainties, offering reliable predictions under diverse observational conditions, whereas SQuAT employs query-driven attention mechanisms to enhance interpretability by explicitly associating spectral features with specific biosignature species. We demonstrate that both models achieve comparably high predictive accuracy on an augmented dataset spanning a wide range of exoplanetary conditions, while highlighting their distinct advantages in uncertainty quantification and spectral interpretability. These capabilities position our methods as promising tools for accelerating target triage, optimizing observation schedules, and maximizing scientific return for upcoming flagship missions such as HWO.
\end{abstract}


% The Emerging Applications track focuses on novel applications of AI methods to real-world problems that are not yet fully deployed. Submissions should demonstrate potential for future deployment, addressing emerging engineering or sociotechnical challenges and their practical relevance. Papers will be evaluated on problem significance, innovation, AI methodology, technical quality, and clarity, with a clear path toward deployment. The Emerging Track is more selective than the Deployed Track, allowing papers that have not yet been fully deployed.

\section{Introduction}
\label{sec:introduction}

The field of exoplanet science is entering a new era of planetary characterization, building on the legacy of \emph{Kepler}, which showed that small, sub-Neptune-size ($R_p \lesssim 4\,R_\oplus$) planets are common \citep{Borucki_2010}. This suggests that Earth-sized, potentially habitable worlds may be frequent, making their detection and study a central priority for future missions. NASA's \textbf{Habitable Worlds Observatory} (HWO) aims to directly image and spectrally characterize many potentially habitable exoplanets in reflected light across the ultraviolet (UV), visible (VIS), and near-infrared (NIR). Because coronagraph observations capture only selected bandpasses and telescope time is scarce, efficient triage and prioritization are essential: a precursor study projected up to two years of observations merely to identify high-priority detailed observation targets \citep{LUVOIR_Study_2018}. With the multi-billion-dollar costs of a flagship like HWO and limited observing time, avoiding even a single 10-hour observation that would not yield useful information represents a near million-dollar cost saving. Advancing these capabilities now will directly inform mission planning and architecture, positioning artificial intelligence (AI) and machine learning (ML) as an operational capability from the outset, initially in ground-based planning and triage, and with a potential path to selective onboard use as flight computing and verification mature.

Direct-imaging pipelines must translate partial spectra into physically meaningful inferences about atmospheres and surfaces (``retrieval''). Classical atmospheric retrieval analysis techniques have emphasized chemical species abundances and surface properties \citep{madhusudhan2018retrieval}, yet abiotic photochemical and geological processes can produce biosignature species leading to false-positive interpretations (e.g., volcanic and hydrothermal activity producing abiotic \ce{CH4}). In contrast, biosignature \emph{fluxes}, or the rates at which biosignatures (e.g., \ce{O2} and \ce{CH4}) enter the environment, tie observations to underlying sources and sinks (i.e., processes that add or remove gases from a planet's atmosphere, respectively) \citep{schwieterman2018exoplanet}. Fluxes that exceed plausible abiotic production offer stronger evidence for life than abundances alone, but retrieving fluxes with traditional approaches is computationally prohibitive: $10^4$–$10^6$ forward-model evaluations combined with minutes-per-simulation photochemistry render end-to-end flux retrieval impractical. Therefore, there is a practical need to accurately infer biosignature fluxes from physically realistic reflected-light spectra in a computationally efficient manner.

Robust uncertainty quantification (UQ) is central to decision-making under low signal-to-noise ratio (SNR), incomplete spectral coverage, and model misspecification. Calibrated posteriors that capture epistemic and aleatoric uncertainty give planners a quantitative basis to balance observing time against expected scientific return, allocating resources only when the probability that a candidate’s gas flux exceeds abiotic limits surpasses a risk-adjusted threshold.

We address these challenges with two complementary models: \textbf{(1) a Bayesian convolutional neural network (BCNN)}, which places UQ at the core of prediction, and \textbf{(2) our novel query-based Transformer, the Spectral Query Adaptive Transformer (SQuAT)}, which employs a compact set of learnable, biosignature-specific query tokens that cross-attend to the wavelength-encoded spectrum using absorption-region priors from the literature and incorporates physics-guided attention priors. Deterministic convolutional neural network (CNN) and vision transformer (ViT) baselines serve as references, and their Bayesian variants propagate epistemic uncertainty via weight distributions and Monte Carlo (MC) dropout. \textbf{SQuAT’s query-driven attention yields interpretable heatmaps and ViT-level accuracy, while the BCNN delivers competitive accuracy with strong uncertainty calibration.} Predictions from both models offer credible intervals for analyzing exoplanets, ranking targets for biosignature detection probability and allocating scarce observation time.

To rigorously train and evaluate these methods, we expand the Frontier Development Lab's (FDL) PyAtmos corpus \citep{chopra2023pyatmos} with non-modern-Earth-like atmospheric grids and pair each self-consistent climate-photochemistry simulation with a reflected-light spectrum generated by the Planetary Spectrum Generator \citep[PSG;][]{villanueva2018planetary}. The resulting dataset spans a wider range of stellar types and surface–atmosphere states.

\section{Related Works}
We find prior literature establishes \textbf{(i)} ML-based retrieval feasibility, \textbf{(ii)} adequate methods for robust uncertainty quantification and interpretability, and \textbf{(iii)} query-specific attention in other domains enabling structured, disentangled outputs \citep{madhusudhan2018retrieval, soboczenski2018inara, yip2020peeking, carion2020detr}. \textbf{However, none integrate all three for biosignature \emph{flux} regression, motivating our unified SQuAT framework.}

% \paragraph{Atmospheric Retrieval and the Move Toward Learning-Based Surrogates.}
% Classical atmospheric retrieval for exoplanets relies on Bayesian sampling over high-dimensional forward models, incurring substantial computational cost. Foundational reviews and frameworks underscore these challenges and motivate faster surrogate approaches \citep{madhusudhan2018retrieval}. Early supervised and ensemble methods (random forests, GANs, and deep neural surrogates) demonstrated order-of-magnitude speedups while maintaining retrieval fidelity, establishing the feasibility of machine learning (ML) accelerators for transmission or emission spectra (e.g., \cite{marquezneila2018supervised, zingales2018exogan, soboczenski2018inara, cobb2019plannet}).

Bayesian deep learning methods, including ensembles and variational or dropout-based approximations, have improved uncertainty calibration and parameter correlations in retrieval tasks \citep{gal2016dropout, soboczenski2018inara, cobb2019plannet}. Studies show that embedding domain structure can enhance both accuracy and interpretability, but focus on static abundance or profile inference, not biosignature flux prediction.

For direct imaging, ML models have enabled rapid estimation of bulk atmospheric properties and triage of terrestrial analogs using albedo or photometric data \citep{johnsen2019mlp, pham2022followwater}. These approaches focus on classification or coarse regression, not multi-output flux prediction.

Interpretability methods reveal spectral models often attend to molecular absorption bands \citep{yip2020peeking, fisher2019highres}. This supports architectures with built-in spectral traceability, such as attention mechanisms aligned with species-specific diagnostics.

Transformers use global self-attention for efficient sequence modeling \citep{vaswani2017attention}. Query-based decoders, such as Detection Transformers (DETR), localize structured outputs via fixed semantic tokens \citep{carion2020detr}. Such designs remain unexplored in exoplanet retrieval, where prior work used shared output heads without task-specific queries.

\section{Dataset}
\label{sec:dataset}

We build on \textbf{PyATMOS}, an open grid of $\sim$125,000 1D coupled photochemistry–climate simulations for an Earth analog around the Sun \citep{chopra2023pyatmos}. Atmospheres with surface temperatures exceeding $\sim$320 K were excluded to conservatively avoid regimes near the inner edge of the habitable zone limit, where runaway greenhouse conditions are expected \citep{Kopparapu_2014}. As a result, we retain \(N_{\text{atm}}\) = 77,882 conservative habitable states.

We extended the dataset to include Proterozoic Earth-like scenarios that capture a broader atmospheric diversity of biologically active planets from an earlier period in Earth's inhabited history. We vary \ce{O2}, \ce{CH4}, and \ce{H2O} on a logarithmic grid from $10^{-15}$ to $10^{-2}$, producing $\sim$1,500 self-consistent equilibrated cases after removing unphysical mixtures (total gas fraction $>1$). The stellar spectrum is scaled for reduced luminosity following \citet{Claire_2012}. Additionally, planetary parameters (gravity, radius, orbital distance) are kept fixed at Earth-like values. The surface pressure is 1~bar, \ce{CO2} and \ce{CO} are fixed at $10^{-2}$ and $10^{-4}$, and \ce{N2} serves as the main background gas.

For non-habitable baselines, we systematically vary the effective stellar flux received by modern Earth by reducing incident stellar flux from 1.0 to 0.3 in steps of 0.02, yielding 36 simulations with subfreezing surfaces.

Each simulation provides altitude-dependent temperature, pressure, and abundances, plus the \emph{net lower-boundary flux} (\(z=0\)) for eight biosignature-relevant gases: \ce{O2}, \ce{O3}, \ce{CH4}, \ce{N2O}, \ce{CO2}, \ce{H2O}, \ce{CO}, \ce{SO2}. These signed fluxes (source $>$0, sink $<$0) are the predictive targets.

We generate reflected-light spectra with PSG using each case’s temperature–pressure and vertical abundance profiles. Spectra span 0.2–2.5~\textmu m (355 points within the UV-VIS-NIR wavelength regions) at resolving power $R=140$, reported as planet–star contrast for an Earth-twin star–orbit configuration. Instrumental noise is added downstream via SNR augmentation (5–100). These spectra serve as inputs to the learning pipeline.

We use 59,202/19,735/19,735 samples for train/val/test and hold out 1,033 TRAPPIST-1e samples, generated with the system’s spectral energy distribution and planetary parameters, for out-of-distribution evaluation.

We apply z-score normalization independently per wavelength using training statistics. Targets \(y\) undergo a per-species \(\operatorname{asinh}\) transform,
\(y'=\operatorname{asinh}(y/\beta)\),
with \(\beta\) set to the training 90th percentile of \(|y|\) for that species, stabilizing heavy-tailed flux distributions.

We intend to open source this augmented dataset in the near future.

\section{Model Architectures}
\label{sec:model_architectures}
In this section, we explain the different model architectures experimented with in our work. Our models take as input raw reflected–light spectra, represented as a single-channel sequence of 355 wavelength points, and the outputs are eight biosignature fluxes, described in detail in Section \ref{sec:dataset}.

For all models, we report point-estimation metrics including the coefficient of determination ($R^{2}$), root-mean-squared error (RMSE), and mean absolute error (MAE); mean squared error (MSE) is used as the loss. For Bayesian or MC-dropout models, we additionally assess predictive uncertainty using coverage-based calibration (fractions within $\pm1\sigma$ and $\pm2\sigma$ of the predictive mean), sharpness (mean predictive standard deviation), the mean coefficient of variation (std$/|\text{mean}|$), and the correlation between predicted uncertainty and absolute error. The final checkpoint is selected via early stopping on validation MSE with a ReduceLROnPlateau scheduler. Model hyperparameters are optimized using a Bayesian algorithm to minimize validation loss, where models presented herein use the best-performing configuration. Development is conducted on a Linux system equipped with a single NVIDIA Tesla V100 GPU (32 GB VRAM) and Intel Xeon Platinum 8270 CPU (2.70GHz). Code was developed using Python 3.11 and PyTorch 2.5, with relevant packages detailed within the repository, which will be open-sourced. Global random seed is set to 42 for reproducibility. 

\subsection{Convolutional and Bayesian Convolutional Neural Networks}
\label{sec:bcnn}
Our baseline is a lightweight 1D CNN. The backbone consists of five Conv1D–ReLU–MaxPool blocks (filters: 32, 64, 128, 256, 512; kernel sizes: 13, 11, 9, 7, 5), followed by two fully connected layers (256 and 128 units, dropout $p_D=0.5$) and a final linear output layer. We train for $\sim$130 epochs using MSE loss with Adam ($\eta=10^{-5}$, batch size 128, dropout $p_D=0.5$). The CNN serves as a lower-bound reference for model performance.

The BCNN retains this architecture but replaces all convolutional and linear layers with Bayesian counterparts that maintain Gaussian weight posteriors~\citep{blundell2015weight}, and uses a heteroscedastic output head to jointly predict the mean and variance for each biosignature~\citep{kendall2017uncertainties}. Training runs for $\sim$140 epochs and minimizes an MSE term plus a Kullback–Leibler (KL) regularizer on the weight posteriors:
\begin{equation}
\mathcal L_{\text{train}}
=\frac{1}{N}\sum_{n=1}^{N}\frac12\lVert y_n-\hat{y}_n\rVert_{2}^{2}
\;+\;
\beta\,\mathrm{KL}\!\bigl(q_{\boldsymbol\theta}(\mathbf W)\,\|\,p(\mathbf W)\bigr),
\end{equation}
where $\beta$ scales the regularization strength. At inference, $T=50$ stochastic forward passes are used to estimate predictive means and decompose total uncertainty into epistemic and aleatoric components.

\subsection{Spectral Query Adaptive Transformer}
\label{sec:SQuAT}
The SQuAT model extends a ViT backbone~\citep{dosovitskiy2020image} to better capture species-specific features in 1D exoplanet spectra. Our ViT configuration uses an embedding dimension $D=256$, $L_{\text{enc}}=6$ Transformer layers, $h=8$ attention heads, MLP ratio $=4$, and dropout $p_D=0.2$. The spectrum is divided into overlapping patches (stride $=1$), embedded with positional encodings, and processed by this stack of multi-head self-attention and feed-forward blocks.

\begin{figure}[htbp]
    \centering
    \includegraphics[width=0.8\columnwidth]{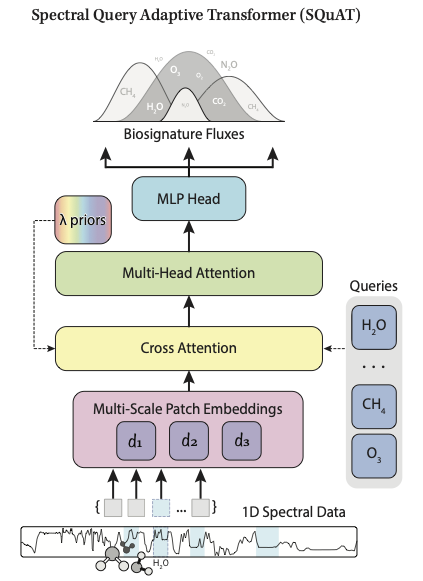}
    \caption{SQuAT model architecture.}
    \label{fig:squat_arch}
\end{figure}

SQuAT augments this design in three key ways. First, it employs a \emph{multi-scale patch encoder} with three parallel convolutional branches (patch sizes 3, 5, and 10) to capture both narrow molecular absorption lines and broader spectral structures. The resulting features are concatenated and projected to a $D=256$-dimensional embedding before entering a Transformer encoder ($L_{\text{enc}}=6$ layers, $h=8$ heads, dropout $p_D=0.2$). Second, instead of a single global output token, SQuAT introduces a set of $K=8$ learnable \emph{biosignature queries}, one per target species. These queries attend to the encoded spectrum via multi-head cross-attention, producing per-species embeddings. This use of a fixed set of biosignature queries is conceptually related to DETR’s “object queries” \citep{carion2020detr}, but SQuAT performs fixed-$K$ per-species flux regression on 1D spectra with physics-guided attention priors and omits Hungarian matching and detection losses. To inject spectroscopic domain knowledge, each attention map is softly biased toward known absorption bands for that species using
\begin{equation}
\mathbf{A'} = (1-\alpha)\,\mathbf{A} + \alpha\,\mathbf{P}
\label{eq:alpha_bias}
\end{equation}
where $\mathbf{A}$ are the learned attention weights, $\mathbf{P}$ is a biosignature-specific prior mask, and $\alpha\in[0,1]$ is a learnable mixing coefficient controlling the strength of prior injection during training. Finally, a \emph{species interaction module} applies self-attention between the species embeddings to capture interdependencies, followed by a lightweight MLP head that produces a scalar flux prediction for each species. A diagram of this architecture is presented in Figure \ref{fig:squat_arch}.

% Put only if space needed (already mentioned in Related Works but always good to explicitly cite an inspiration)
% This use of a fixed set of biosignature queries is conceptually related to DETR’s “object queries” \citep{carion2020detr}, but SQuAT performs fixed-$K$ per-species flux regression on 1D spectra with physics-guided attention priors and omits Hungarian matching and detection losses.

Predictive uncertainty is estimated using MC dropout~\citep{gal2016dropout}: dropout remains active at inference and $T=30$ stochastic forward passes yield the predictive mean and epistemic variance. SQuAT is trained for $\sim$35 epochs, while the ViT baseline is trained for $\sim$50 epochs, and both use Adam ($\eta=10^{-4}$, batch size 64).

\section{Results}
\label{sec:results}

\begin{figure}[htbp]
    \centering
    \includegraphics[width=\columnwidth]{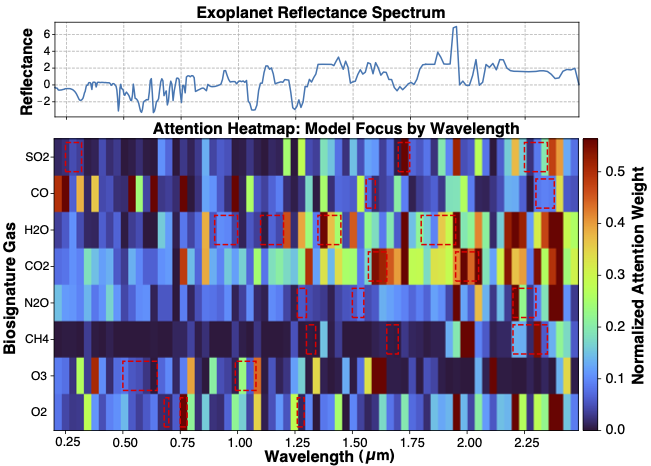}
    \caption{SQuAT attention map. Top: reflected-light exoplanet spectrum sample. Bottom: per-species attention across wavelength, highlighting prominent regions for each biosignature’s flux prediction. Red dashed boxes mark canonical absorption bands. Attention is normalized per species.}
    \label{fig:attention_map}
\end{figure}

Figure \ref{fig:attention_map} visualizes the attention weights of one exoplanet sample learned by the SQuAT model across the input wavelength range. Each row corresponds to a target biosignature, with color intensity reflecting the relative importance assigned to each spectral region. Normalizing attention weights independently per biosignature enables comparison of intra-species focus without being dominated by stronger species. Attention weight mapping enhances diagnostic analysis and transparency by revealing which parts of the spectrum the model attends to when making predictions.

\begin{figure}[htbp]
    \centering
    \includegraphics[width=0.85\columnwidth]{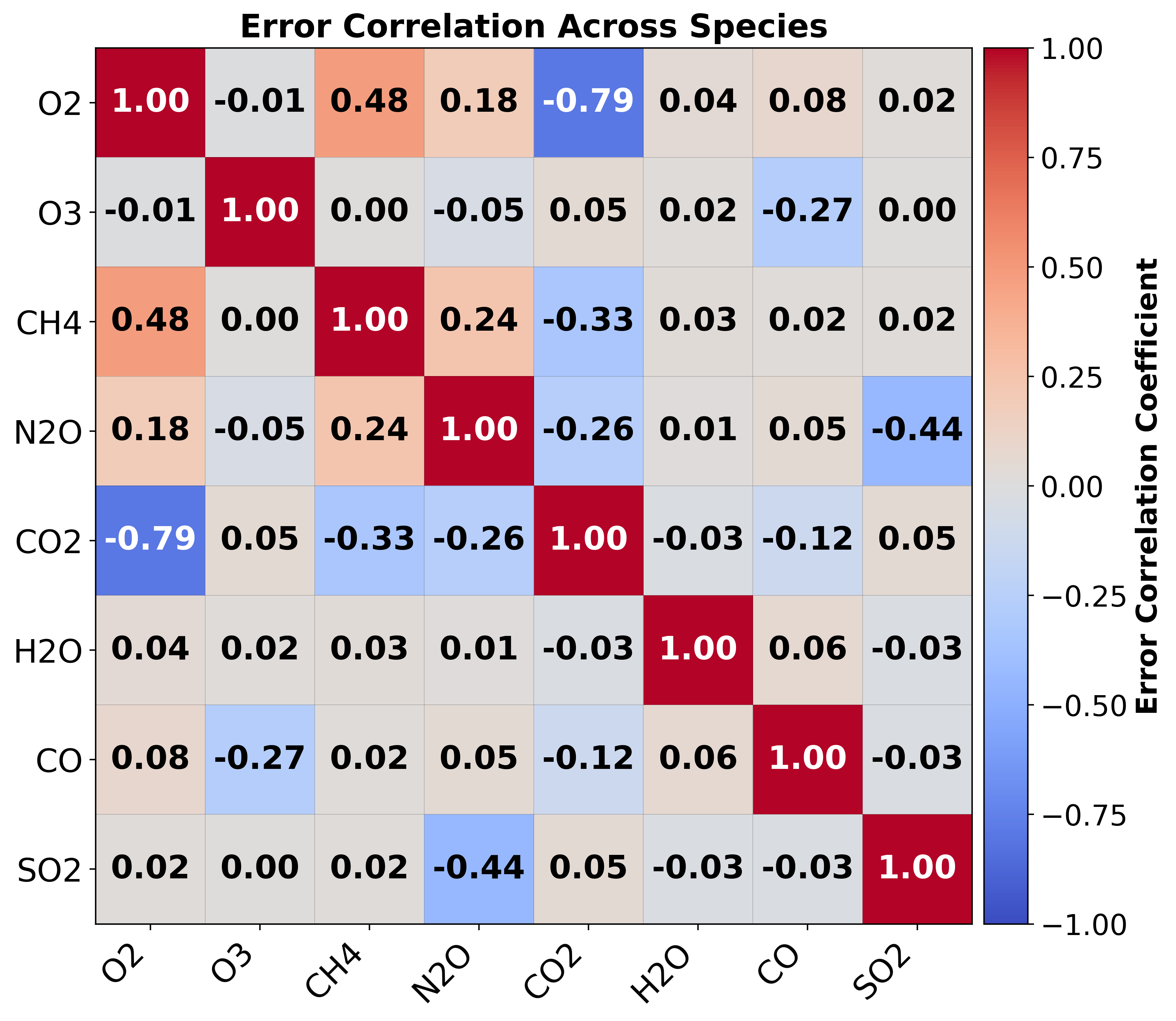}
    \caption{Error correlation matrix across predicted biosignature species. Each cell shows the Pearson correlation coefficient between the model prediction errors for a pair of species. High positive values (red) indicate that errors tend to co-occur in the same direction, while strong negative values (blue) suggest opposing error trends.}
    \label{fig:error_correlation_heatmap}
\end{figure}

Figure \ref{fig:error_correlation_heatmap} describes the Pearson correlation coefficients between prediction errors across all modeled species. Each matrix entry quantifies the degree to which errors in one species’ predicted flux are linearly associated with errors in another. Notably, errors in predicting \ce{O2} and \ce{CO2} are strongly anti-correlated ($r = -0.79$), suggesting shared spectral or model confusion. This analysis reveals potential dependencies or disentanglements in the model’s learned representations across atmospheric species.

\begin{figure*}[t]
  \centering
  \begin{minipage}[t]{0.5\textwidth}
    \centering
    \includegraphics[width=\linewidth]{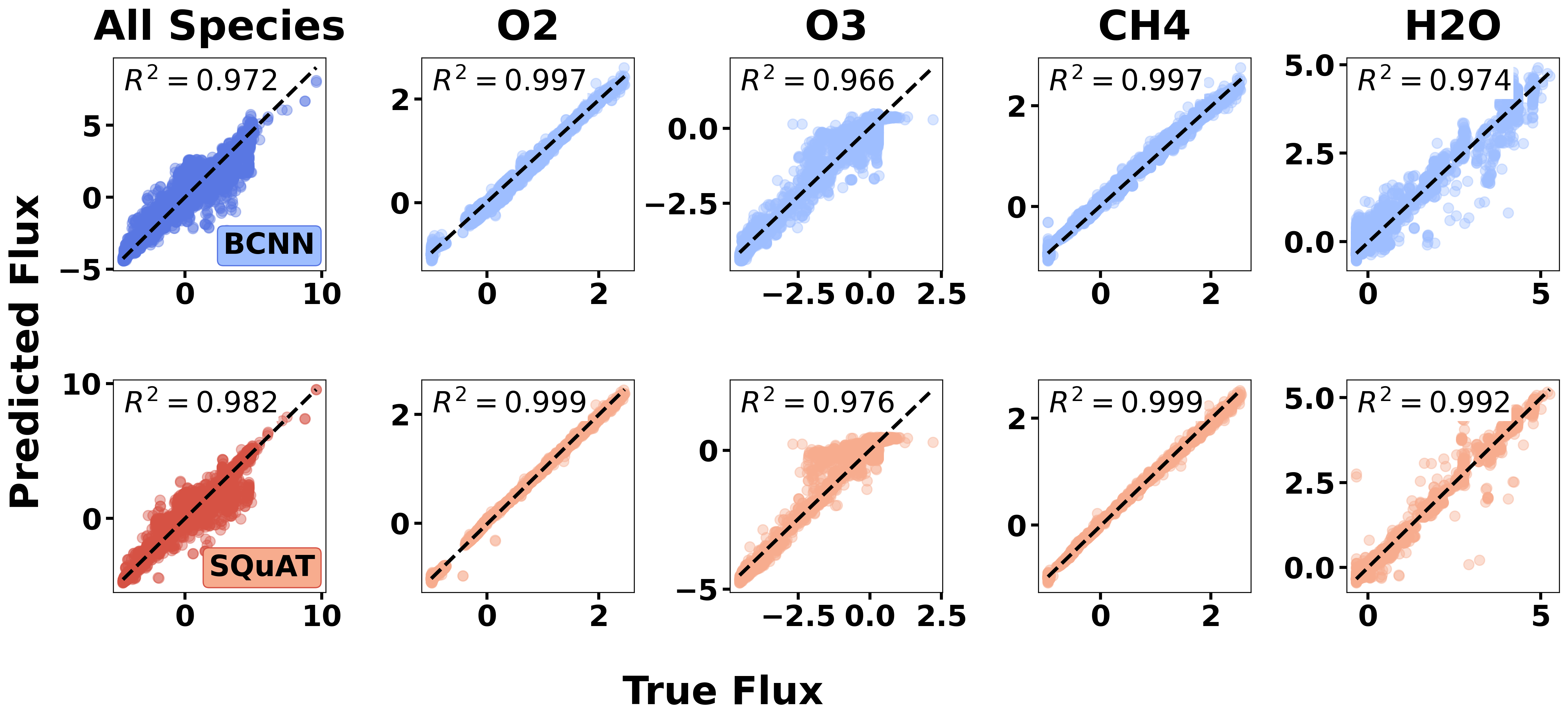}\\[2pt]
    \textbf{(a)}
  \end{minipage}\hfill
  \begin{minipage}[t]{0.5\textwidth}
    \centering
    \includegraphics[width=\linewidth]{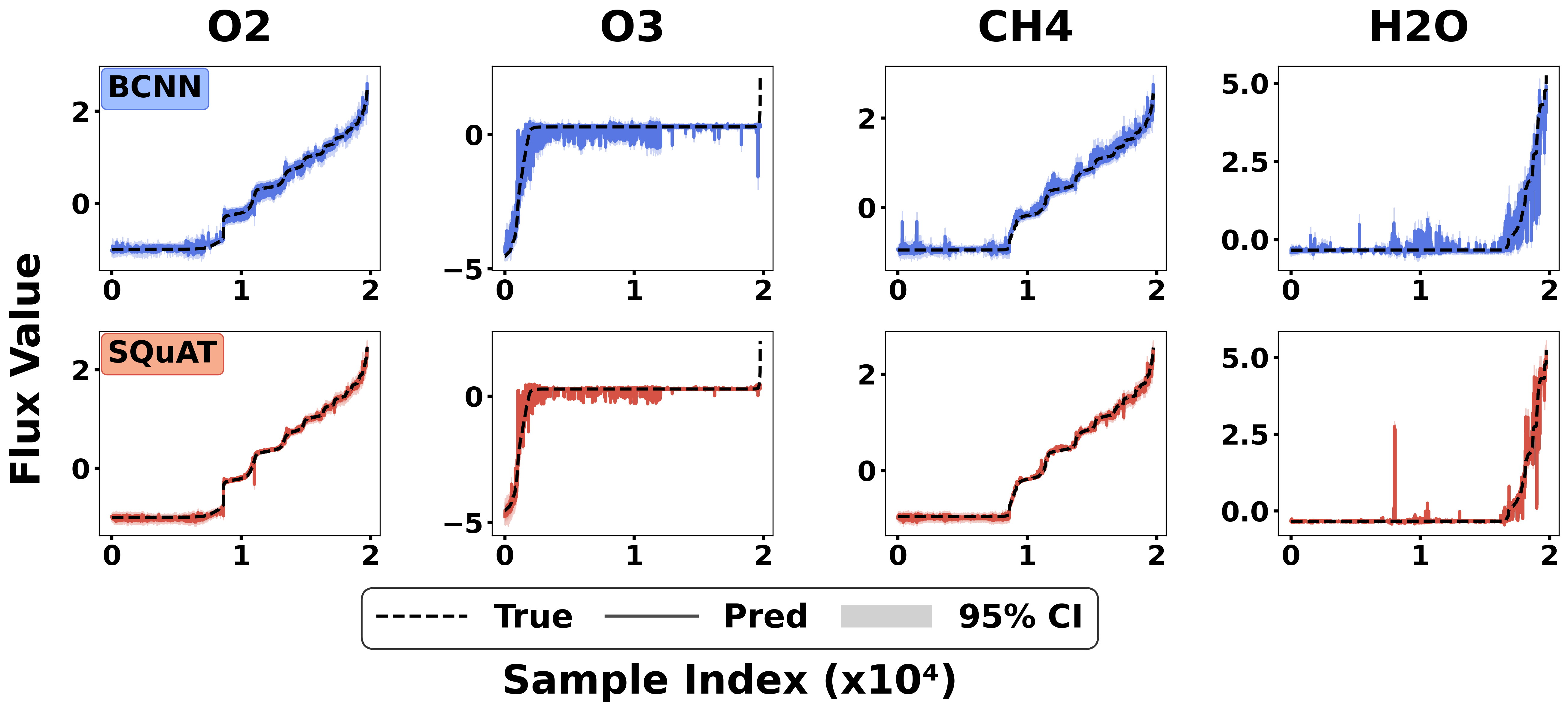}\\[2pt]
    \textbf{(b)}
  \end{minipage}
  \caption{(a) Predicted versus true biosignature flux for four molecular species (\ce{O2}, \ce{O3}, \ce{CH4}, \ce{H2O}) and for all species combined. (b) Corresponding model predictions with 95\% credible intervals, with samples sorted by increasing true flux.
  }
  \label{fig:scatter_uncertainty_sidebyside}
\end{figure*}

Figure~\ref{fig:scatter_uncertainty_sidebyside}a presents scatter plots of predicted versus true biosignature flux values for four selected species—\ce{O2}, \ce{O3}, \ce{CH4}, and \ce{H2O}—as well as for all species aggregated. The top row depicts results from the BCNN, while the bottom row shows results from the SQuAT. For each model, we compute $R^2$ as a measure of fit quality. Across all targets, SQuAT achieves higher $R^2$ scores, indicating stronger agreement with ground truth. The dotted diagonal line represents perfect prediction.

Figure~\ref{fig:scatter_uncertainty_sidebyside}b displays predicted flux values with associated 95\% credible intervals for each of four molecular species: \ce{O2}, \ce{O3}, \ce{CH4}, and \ce{H2O}. Each subplot shows the true values sorted in ascending order on the x-axis, allowing for clearer visualization of prediction fidelity and uncertainty across the value range. The BCNN (top row) and the SQuAT (bottom row) both achieve high $R^2$ scores, indicating accurate performance. Notably, both models capture structured uncertainty that varies with flux magnitude, and visually track epistemic variability, particularly for more challenging species like \ce{O3}. Credible intervals are narrowest for well-predicted regions and broaden with greater flux variability.

\begin{table}[t]
    \centering
    \setlength{\tabcolsep}{1mm} % tighter columns for readability
    \renewcommand{\arraystretch}{1.15} % a bit more row height
    \resizebox{\columnwidth}{!}{
    \begin{tabular}{c|cc|cc|cc|cc}

\multicolumn{1}{l}{\textbf{}} & \multicolumn{2}{c}{\textbf{CNN}} & \multicolumn{2}{c}{\textbf{BCNN}} & \multicolumn{2}{c}{\textbf{ViT}} & \multicolumn{2}{c}{\textbf{SQuAT}} \\ \hline
\textbf{SNR}                   & $R^2$           & RMSE            & $R^2$            & RMSE            & $R^2$           & RMSE            & $R^2$            & RMSE            \\ \hline
5                              & 0.155           & 0.917           & 0.812            & 0.432           & 0.827           & 0.415           & 0.747            & 0.502           \\
10                             & 0.806           & 0.439           & 0.933            & 0.258           & 0.939           & 0.247           & 0.924            & 0.274           \\
20                             & 0.939           & 0.246           & 0.962            & 0.195           & 0.968           & 0.179           & 0.970            & 0.174           \\
40                             & 0.966           & 0.184           & 0.970            & 0.174           & 0.982           & 0.136           & 0.979            & 0.144           \\
50                             & 0.969           & 0.175           & 0.971            & 0.171           & 0.983           & 0.131           & 0.981            & 0.139           \\
100                            & 0.973           & 0.164           & 0.972            & 0.167           & 0.985           & 0.122           & 0.982            & 0.134           \\ \hline
\end{tabular}
    }
    \caption{$R^2$ and RMSE for models at selected target SNRs.}
    \label{table:r2_rmse}
\end{table}

Table~\ref{table:r2_rmse} reports $R^2$ and RMSE for CNN, BCNN, ViT, and SQuAT across target SNRs (5–100). Scores improve with SNR for all models. For SNR $\geq 20$, ViT and SQuAT differ by at most $\Delta R^2\leq 0.003$ and $\Delta\text{RMSE}\leq 0.013$ (e.g., at SNR$=20$: $R^2=0.970$ vs.\ $0.968$, RMSE$=0.174$ vs.\ $0.179$), whereas at SNR$=5$ the spread is larger (e.g., ViT $0.827/0.415$ vs.\ SQuAT $0.747/0.502$; CNN $0.155/0.917$).

% replace above paragraph with this if you need space
% Table~\ref{table:r2_rmse} summarizes $R^2$/RMSE across SNR 5–100; all models improve with SNR. For SNR$\geq20$, ViT and SQuAT differ by at most $\Delta R^2\leq0.003$ and $\Delta\mathrm{RMSE}\leq0.013$ (SNR$=20$: $0.970/0.174$ vs.\ $0.968/0.179$), whereas at SNR$=5$ the spread is larger (ViT $0.827/0.415$, SQuAT $0.747/0.502$, CNN $0.155/0.917$).

\section{Discussion}
\label{sec:discussion}

To assess whether the model has learned scientifically meaningful associations between spectral features and biosignature species, we visualize the learned attention weights across the input spectrum for a single representative sample (Figure~\ref{fig:attention_map}). Encouragingly, the model attends strongly to wavelength regions that correspond to known molecular absorption features for most species. For instance, \ce{H2O} attention peaks align with established bands at 1.1~\textmu m, 1.4~\textmu m, and $\sim$1.9~\textmu m~\citep{rothman1998hitran}. \ce{CH4} exhibits clear attention near its strong absorption window at 2.3~\textmu m, while \ce{CO2} focuses on features near 2.0~\textmu m. Importantly, the model successfully captures the \ce{O2} 0.76~\textmu m A-band, which is one of the strongest, most accessible oxygen features for direct imaging; detection of a pronounced A-band can indicate substantial atmospheric \ce{O2}, which on rocky planets is often associated with oxygenic photosynthesis and potential biological activity~\citep{schwieterman2018exoplanet}. However, the \ce{O2} 1.27~\textmu m band receives only modest attention, while that same region shows relatively strong attention for \ce{CO2}, which may contribute to their observed error correlation (Figure~\ref{fig:error_correlation_heatmap}). Interestingly, this is consistent with how retrieval methods like that used for the Total Carbon Column Observatory Network use the 1.27~\textmu m \ce{O2} band as a proxy for total air column when inferring \ce{CO2} columns~\citep{Bertaux2020_TCCON_O2proxy}, suggesting that while the model did not successfully predict \ce{O2} flux, it learned other scientifically relevant relationships embedded in this spectral region, such as its role as a proxy for total atmospheric column, which can indirectly inform \ce{CO2} predictions. We also note partially overlapping short-wave infrared attention for \ce{CH4} and \ce{CO2} in the 2.0--2.3~\textmu m range, which may also contribute to correlated prediction errors (Figure~\ref{fig:error_correlation_heatmap}), though \ce{CO2}'s strongest sensitivity remains closer to 2.06~\textmu m. Finally, we observe attention patterns outside the conventionally highlighted absorption bands, suggesting either error in the model, or identification of additional subtle or blended spectral features that contribute to its predictions. It is important to note that Figure~\ref{fig:attention_map} reflects the attention pattern for a single spectrum; if the model pays little attention to a canonical absorption region for a given molecule, it may simply indicate that the corresponding species is absent or weak in that specific spectrum.

Figure~\ref{fig:scatter_uncertainty_sidebyside}a and Table~\ref{table:r2_rmse} show high $R^2$ across species for both BCNN and SQuAT. This close performance ($R^2 = 0.972$ for BCNN and $R^2 = 0.985$ for SQuAT) on all-species regression suggests that both models are near the representational ceiling given the SNR ratio and data complexity. On all-species regression SQuAT is only marginally above BCNN ($0.985$ vs.\ $0.972$), and gaps are small for SNR$\geq20$. By design, these variants were not intended to raise point accuracy but to add capabilities. BCNN replaces deterministic layers with variational ones to quantify predictive uncertainty, and SQuAT augments a ViT with gas-query cross-attention and spectroscopic priors to expose species-aligned attributions. Consistent with this intent, their point accuracy largely tracks the CNN/ViT baselines; the added value is complementary to accuracy. BCNN provides predictive distributions with credible intervals for risk-aware use, and SQuAT yields attention maps that localize the wavelengths driving each prediction. Thus, even when $R^2$ and RMSE are nearly indistinguishable, the models supply the quantified uncertainty and interpretable attributions needed for traceable flux retrievals and observation planning.

% Uncertainty bands figure
The shape and magnitude of the 95\% credible intervals in Figure~\ref{fig:scatter_uncertainty_sidebyside}b provide insight into each model’s internal uncertainty calibration. While both models show narrow bands for well-constrained species like \ce{O2} and \ce{CH4}, the SQuAT appears to express sharper and more localized uncertainty, potentially reflecting its attention-based structure and discrete focus on relevant wavelengths. Conversely, the BCNN shows smoother uncertainty profiles that may stem from the global nature of convolutional filtering and weight-level variational inference. These patterns suggest transformer-based models might be better suited for uncertainty disentanglement in spectrally entangled regimes, though further calibration metrics would be needed to quantify this.

Our results highlight distinct strengths of the BCNN and the SQuAT, each making them particularly suitable for different operational and scientific priorities in exoplanet biosignature flux prediction. The BCNN exhibits notable performance in terms of robust uncertainty quantification, which is crucial in scenarios demanding reliability under varied observational conditions. Its probabilistic nature provides inherent flexibility to capture both aleatoric and epistemic uncertainties. Thus, the BCNN is particularly well-suited to mission planning phases where conservative decision-making is necessary. For instance, it can significantly contribute to preliminary prioritization of observational targets by efficiently identifying planetary atmospheres with the greatest scientific potential, even under conditions of limited data quality or incomplete spectral coverage. The smoother uncertainty profile generated by the BCNN (as observed in Figure~\ref{fig:scatter_uncertainty_sidebyside}b) also suggests potential advantages for broader generalization across observational regimes with varying noise levels.

In contrast, the SQuAT leverages a query-driven attention mechanism to explicitly disentangle molecule-specific spectral features. This architectural choice enhances interpretability, allowing scientists to directly associate predictions with known molecular absorption bands. SQuAT’s structured attention mechanism enables focused extraction of spectral signals, making it particularly suitable for detailed scientific analysis where understanding the specific spectral drivers behind predictions is essential. The ability to visualize per-molecule attention maps (as demonstrated in Figure~\ref{fig:attention_map}) provides valuable insight into the underlying physics captured by the model, making SQuAT an excellent candidate for exploratory data analysis and hypothesis generation. Furthermore, SQuAT’s multi-scale processing and explicit incorporation of spectroscopic priors enhance its robustness to spectral overlap between molecules, increasing confidence in disentangled flux estimations.

Interestingly, despite these differences in architecture and interpretability, both models achieve comparable and exceptionally high predictive performance (Figure~\ref{fig:scatter_uncertainty_sidebyside}a and Table \ref{table:r2_rmse}). This observation underscores an important point: while architectural sophistication (as exemplified by SQuAT) provides significant interpretative and analytical advantages, the simpler BCNN remains highly effective and may represent a preferable choice when computational resources are constrained, or when interpretability is not the priority. \textbf{Ultimately, the complementary strengths of SQuAT and BCNN suggest an ensemble approach may address the dual requirements of future flagship missions seeking signs of habitability}.

\section{Limitations and Future Work}
\label{sec:future_work}

One main limitation for our models is the lack of real-world observations. Presently, the vast majority of exoplanet spectra come from the transit method, which is mainly successful on planets significantly hotter than Earth and unlikely to contain life as we know it \cite{stark2020toward}. Additionally, simulated data have their own limitations and biases. For example, the Atmos simulation described in Section \ref{sec:dataset} does not properly converge at higher temperature regimes \cite{vpl_atmos}, and clouds and hazes remain challenging to model realistically.

The scarcity of observational data necessitates comprehensive synthetic datasets. We will augment the dataset to include a broader diversity of exoplanetary atmospheres and incorporate emerging simulation capabilities as they become available. A larger dataset also enables development of a pre-trained foundation model, with biosignature flux prediction as a downstream fine-tuning task.

Increasing model robustness at low SNR is a key priority, because practical applications must assume minimal signal quality. We will explore other Bayesian deep learning approaches, integrate additional physics priors into learned representations, and develop attention mechanisms that better capture localized spectral dependencies while disentangling overlapping molecular features.
\section{Path to Deployment}
\label{sec:path_to_deployment}
The Habitable Worlds Observatory (HWO) is envisioned as NASA’s next flagship direct-imaging mission, designed to detect and characterize Earth-sized exoplanets in the habitable zones of nearby stars  \cite{zurlo2024direct}. With its enormous cost, finite mission lifetime, and extremely limited observing-time budget, HWO must maximize scientific return from every observation. An AI-assisted decision pipeline offers a compelling route to achieve this, and our work represents a critical step toward this by demonstrating high-accuracy, uncertainty-aware biosignature flux inference from realistic reflected-light spectra. Together, our models satisfy two mission-critical AI requirements: risk-aware decision-making and scientifically traceable predictions.

To reach deployment, milestones are structured according to the \textbf{Technology Readiness Level} (TRL) framework, a systematic metric used by NASA and other space agencies to assess status and operational readiness of mission technologies \cite{NASA_SEH_2016}. Progress to lower-mid TRL includes advancing our models with increasingly realistic synthetic spectra training data covering diverse planetary atmospheres, surface types, clouds, and hazes. These are fine-tuned on HWO-specific end-to-end simulations to incorporate realistic noise, coronagraph bandpasses, and throughput models, ensuring hardware-resilient algorithms. To reach higher TRL, simulations would quantify how AI-guided prioritization improves science yield under realistic operational constraints like slew times and visibility windows. Once validated, models can be deployed on the ground during commissioning and early science operations, serving as decision-support tools alongside human experts. Upon achieving highest TRL, models can operate within the Science Operations Center (SOC) for real-time observation triage.
\section{Conclusion}
\label{sec:conclusion}
In this study, we introduced and evaluated two deep-learning approaches for predicting exoplanet biosignature fluxes: a BCNN and SQuAT. Both address key requirements of future direct-imaging missions, including robust uncertainty quantification and interpretability. The BCNN excelled at capturing epistemic and aleatoric uncertainties, ideal for limited data scenarios. SQuAT, our novel model architecture, provided superior interpretability through explicit attention mechanisms aligned with biosignature absorption priors, facilitating exploratory analyses and hypothesis validation.

Both models exhibited comparably high predictive accuracy, reinforcing their potential as complementary components within a unified analytical pipeline. Identified limitations, such as spectral confusion between different biosignatures, highlight areas for improvement and establish a need for broader observational scenarios and ensemble methods that combine Bayesian inference with query-based transformers. Overall, our findings highlight the significant promise and flexibility of machine-learning approaches in the search for habitable worlds and their role in maximizing scientific yield from upcoming flagship missions.

%%%% ACKNOWLEDGEMENTS SECTION OMITTED FROM INITIAL SUBMISSION %%%%
% \section{Acknowledgments}
% \label{sec:acknowledgements}

\bibliography{aaai2026}

\end{document}